\pdfoutput=1

\documentclass[11pt]{article}

\usepackage[]{acl}

\usepackage{times}
\usepackage{latexsym}
\usepackage{graphicx}
\usepackage{arydshln}

\usepackage[T1]{fontenc}

\usepackage[utf8]{inputenc}

\usepackage{microtype}

\title{drsphelps at SemEval-2022 Task 2: Learning idiom representations using BERTRAM}

\author{Dylan Phelps \\
  Healthy Lifespan Institute \\
  Department of Computer Science, The University of Sheffield\\
  Sheffield, United Kingdom \\
  \texttt{drsphelps1@sheffield.ac.uk}}

\begin{document}
\maketitle
\begin{abstract}
This paper describes our system for SemEval-2022 Task 2 Multilingual Idiomaticity Detection and Sentence Embedding sub-task B. We modify a standard BERT sentence transformer by adding embeddings for each idiom, which are created using BERTRAM and a small number of contexts. We show that this technique increases the quality of idiom representations and leads to better performance on the task. We also perform analysis on our final results and show that the quality of the produced idiom embeddings is highly sensitive to the quality of the input contexts.
\end{abstract}

\section{Introduction}


Idiomatic expressions present a challenge to Large Language Models (LLMs) as their meaning cannot necessarily be derived from the composition of their component tokens, a trait that LLMs often exploit to create representations of multi-word expressions. The lack of compositionality leads to poor representations for idiomatic expressions and in turn poor performance in downstream tasks whose data includes them.

SemEval-2022 task 2b \citep{tayyarmadabushi-etal-2022-semeval} encourages the creation of better representations of idiomatic expressions across multiple languages by presenting a \textbf{Semantic Text Similarity (STS)} task in which correct STS scores are required whether or not either sentence contains an idiomatic expression. The sub-task requires the creation of a self-consistent model in which a sentence including an idiomatic expression and one containing its literal meaning ('\textit{swan song}' and '\textit{final performance}') are exactly similar to each other and equally similar to any other sentence.

To achieve this goal, we investigate whether due to the similarity between idioms and rare-words Schick and Sch\"utze's BERT for Attentive Mimicking \citep{schick-schutze-2020-bertram} (BERTRAM) model, which was designed for use with rare-words, can be used to explicitly learn high-quality embeddings for idiomatic expressions. We also investigate how many examples of each idiom are required to create embeddings that perform well on the task, as well as how the quality of contexts fed to the BERTRAM model effects the representations and performance on the task.

Evaluating our model on the task shows that externally trained idiom embeddings significantly increase the performance on STS data containing idioms while maintaining high performance on general STS data. This improved performance gained an overall spearman rank score of 0.6402 and first place (of six entries) on the pre-train setting, and an overall spearman rank score of 0.6504 and second place (of five entries) on the fine-tune setting.\footnote{The code for creating the embeddings and the modified baseline system code can be found on GitHub: https://github.com/drsphelps/semeval-task-2.}

\section{Background}
\begin{table*}[h!]
\small
\centering
\begin{tabular}{| p{0.2\linewidth} | p{0.7\linewidth} |}
\hline
\textbf{Usage} & \textbf{Example in Sentence} \\
\hline
\hline
Idiomatic & Blockchains, fundamentally, are banking because what they’re doing is allowing the transaction of value across networks … they’re doing it in an orthogonally different way," he said Wednesday in what may be his \textbf{swan song} in public office. \\
\hline
Literal & Blockchains, fundamentally, are banking because what they’re doing is allowing the transaction of value across networks … they’re doing it in an orthogonally different way," he said Wednesday in what may be his \textbf{bird song} in public office. \\
\hline
Semantically Similar & Blockchains, fundamentally, are banking because what they’re doing is allowing the transaction of value across networks … they’re doing it in an orthogonally different way," he said Wednesday in what may be his \textbf{final performance} in public office. \\
\hline
\end{tabular}
\caption{\label{data-example}
Example sentences for the Idiomatic STS data. Idiomatic and Semantically similar should be given an STS score of 1, and be given the same score when compared to the literal use.
}
\end{table*}

Adopting the idiom principle \citep{sincliar:1991} to produce a single token representation for MWEs has been used widely within static embedding distributional semantic models (\citealp{Mikolov2013DistributedRO}; \citealp{cordeiro-etal-2019-unsupervised}). Within contextualised representation models, \citealp{hashempour-villavicencio-2020-leveraging} show that the contextualised representations produced by context2vec \citep{melamud-etal-2016-context2vec} and BERT \citep{devlin-etal-2019-bert} models can be used to differentiate between idiomatic and literal uses of MWEs. However, the MWEs are only represented by one token in the input, before being broken into many tokens using BERTs word piece tokenizer. \citealp{tayyar-madabushi-etal-2021-astitchinlanguagemodels-dataset} add a token to the BERT embedding matrix and shows that this method improves representations through increased performance on their proposed STS task. The embeddings they add to BERT are randomly initialised, however, and only trained during the fine-tun step on limited data.

\subsection{BERTRAM}

BERT for Attentive Mimicking (BERTRAM) \citep{schick-schutze-2020-bertram}, originally developed to improve representations of rare words, builds upon attentive mimicking \citep{schick-schutze-2019-attentive} to create embeddings, within existing embedding spaces, for tokens that incorporate both form and context information from a small number of example contexts. During training the model attempt to recreate embeddings for common words with the existing embedding in the model treated as the `gold embedding', a process known as mimicking. Form embeddings are then learnt using trained n-gram character embeddings, before being passed with a context into a BERT model. The output of the BERT model forms the embedding for that specific context. To incorporate knowledge from many contexts an attention layer is applied over the outputs for each context to get the final embedding. There exist other models to produce effective embeddings from a small number of contexts \citep{zhao-etal-2018-generalizing, pinter-etal-2017-mimicking}, however, BERTRAM is the only model that is non-bag-of-words and incorporates both form and context information when creating the embedding.

Rare words are unsurprisingly defined by how uncommon they are within datasets. This leads to problems when using LLMs on tasks involving rare words as the word pieces they are broken down into have not been influenced enough during pre-training to accurately represent them. Similarly, idiomatic phrases represent a small proportion of the usage of their constituent words, the idioms in the development set for this task represent an average of 4.9\% of the usage of their constituent words. Therefore, the embeddings for constituent words are not significantly effected by the usage of idioms in the training data, leading to the model failing to understand the idiomatic expressions. Further similarities between idioms and rare-words include the variance in compositionality, for example, \textit{unicycle} can be partially understood from its word pieces, whereas \textit{kumquat} cannot.


\section{Methodology}
\subsection{Embedding Creation}
Due to the similarities between rare words and idioms, we use BERTRAM to create representations for idiomatic expressions. A separate BERTRAM model is used for each nof the tasks languages. For English, we use the pre-trained model provided with the original paper. For Portuguese and Galician we train BERTRAM models with BERTimbau Base \citep{portugueseBERT} and Bertinho-Base \citep{galicianBERT} respectively used as the base transformers. The Portuguese and Galician BERTRAM models that we train are trained using almost the same training regime outlined for the English model in the original paper, 3 epochs of context only training, 10 epochs of form only training and 3 epochs of combined training. Due to time and compute restrictions, we do not use One-Token Approximation to expand the number of gold standard representations that can be used for attentive mimicking. The Portuguese and Galician splits of the cc100 dataset (\citealp{conneau-etal-2020-unsupervised}; \citealp{wenzek-etal-2020-ccnet}) are used to train the models, with the entire split being used for Galician, and a 10GB subset used for Portuguese.

Contexts for each of the idioms found in the task data can then be created using these models. Examples are retrieved from the relevant split in the cc100 dataset using a grep command \footnote{\textit{grep -i " \$val" -m250 en.txt > \$val.data}, where \$val is the idiom of interest} that retrieves the entire line that the instance of the idiom is found on. We investigate how changing the number of contexts used to create each embeddings changes our performance on the task by creating embeddings for each idiom with between 1-250 examples in intervals.

\subsection{Model Architecture}

\begin{figure}
  \centering
  \includegraphics[width=2.7in]{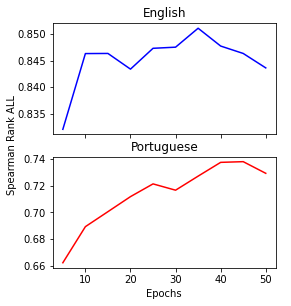}
  \caption{Overall Spearman Rank performance on the development set for the English and Portuguese models at different epochs during pretraining}
  \label{fig:pretrain}
\end{figure}

\begin{figure}
  \centering
  \includegraphics[width=2.7in]{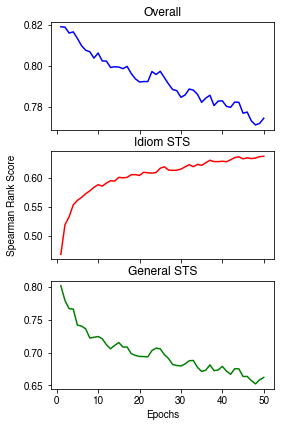}
  \caption{Overall and Idiom STS Only Spearman Rank on the development set whilst training on the Idiom STS data}
  \label{fig:finetune}
\end{figure}

For predicting the similarity scores, a separate model is used for each of the languages BERT-Base \citep{devlin-etal-2019-bert} for English, BERTimbau for Portuguese, and Bertinho-Base for Galician. The created BERTRAM embeddings for each of the idioms found within the task are added into the embedding matrix of the relevant model. These models are used within a Sentence BERT \citep{reimers-gurevych-2019-sentence} setup, implemented using the SentenceTransformers library, which consists of a siamese network structure that uses mean squared error over the cosine similarities of the input sentences as it's loss function. This allows us to use the contextualised embedding outputs of our BERT networks to find cosine similarity between a given pair of sentences.

%


\subsection{Data}

This sub-task uses data in English, Portuguese and Galician. Data is also split into general STS data which does not necessarily contain idioms and idiom STS data which specifically contains idioms and phrases which are semantically similar or literally similar. An example of idiom STS data taken from the task description can be seen in Table \ref{data-example}.

 English and Portuguese are the primary languages and general STS data, from STSBenchmark \citep{cer-etal-2017-semeval} and ASSIN2 \citep{Real2020TheA2} for English and Portuguese respectively, and idiom STS data for both languages are included in the train, dev, eval and test sets. A very small amount (50 examples) of Galician data, comprised of idiom STS data, is also included in the test set. 
 
 The task is split into two settings, pre-train and fine-tune. The pre-train setting does not allow for the use of STS score annotated data which includes idioms, whereas any data can be used in the fine-tune setting.

The evaluation metric used in this task is the correlation between the predicted similarities and the gold standard ones, calculated using Spearman's Rank Correlation Coefficient. The Spearman's Rank is calculated for the general STS data and the idiom STS data separately, however, the Spearman's Rank for the entire dataset is used in the final evaluation.

\subsection{Pre-train Setting}

For the pre-train setting, we use the general STS data in English and Portuguese to train the respective models. Due to a lack of available STS data for Galician, it is trained on the Portuguese data, as there is a high level of similarity between the two languages.

Evaluating the models on the dev split, we investigate the optimal number of epochs for the English and Portuguese models. The results (shown in figure \ref{fig:pretrain}) show that 45 epochs are optimal for Portuguese and 35 for English. Due to a lack of dev split data for Galician we use the result from the Portuguese model as they are trained on the same data.

\subsection{Fine-tune Setting}
\begin{table*}
\centering
\begin{tabular}{llrrr}
\hline
Setting & Language(s) & SR ALL & SR Idiom & SR STS \\
\hline
Pre-Train & EN & 0.7445 & 0.4422 & 0.8709 \\
Pre-Train & PT & 0.7087 & 0.4806 & 0.8010 \\
Pre-Train & GL & 0.2924 & 0.2924 & - \\
\textbf{Pre-Train} & \textbf{All} & \textbf{0.6402} & \textbf{0.4030} & \textbf{0.8641} \\
\textit{Pre-Train} & \textit{EN} & \textit{0.5958} & \textit{0.2488} & \textit{0.8300} \\
\textit{Pre-Train} & \textit{PT} & \textit{0.5584} & \textit{0.2761} & \textit{0.7745} \\
\textit{Pre-Train} & \textit{GL} & \textit{0.1976} & \textit{0.1976} & \textit{-} \\
\textit{Pre-Train} & \textit{All} & \textit{0.4810} & \textit{0.2263} & \textit{0.8311} \\
\hdashline
Fine-Tune & EN & 0.7643 & 0.4861 & 0.8344 \\
Fine-Tune & PT & 0.7307 & 0.4643 & 0.7908 \\
Fine-Tune & GL & 0.2859 & 0.2859 & - \\
\textbf{Fine-Tune} & \textbf{All} & \textbf{0.6504} & \textbf{0.4124} & \textbf{0.8188} \\
\textit{Fine-Tune} & \textit{EN} & \textit{0.6684} & \textit{0.4109} & \textit{0.6210} \\
\textit{Fine-Tune} & \textit{PT} & \textit{0.6026} & \textit{0.4090} & \textit{0.5523} \\
\textit{Fine-Tune} & \textit{GL} & \textit{0.3842} & \textit{0.3842} & \textit{-} \\

\textit{Fine-Tune} & \textit{All} & \textit{0.5951} & \textit{0.3990} & \textit{0.5961} \\
\hline
\end{tabular}
\caption{\label{final-results}
Final Spearman Rank (SR) scores of the system on the test set, split into idiom Semantic Text Similarity (STS), general STS, and all datasets. Aggregated results for all languages in bold. Results for the baseline system, also broken down into languages, are in italics.
}
\end{table*}

For the fine-tune setting we start with the models from the pre-train setting, and further train them on the Idiom STS data provided as part of the task.

Again we investigate the optimal number of epochs of training on this data (results shown in figure \ref{fig:finetune}). We find that the overall spearman rank is highest after just a single epoch of training, with further training considerably reducing the performance on the general STS data, and thus on the overall STS score. However, further training, up to 50 epochs, continues to increase the performance of the model on Idiom STS data. Therefore, depending on the application and required trade-off, the model can be tuned to either perform better on general STS data or idiom STS data.

\subsection{Number of Examples}
\begin{figure}
  \centering
  \includegraphics[width=3in]{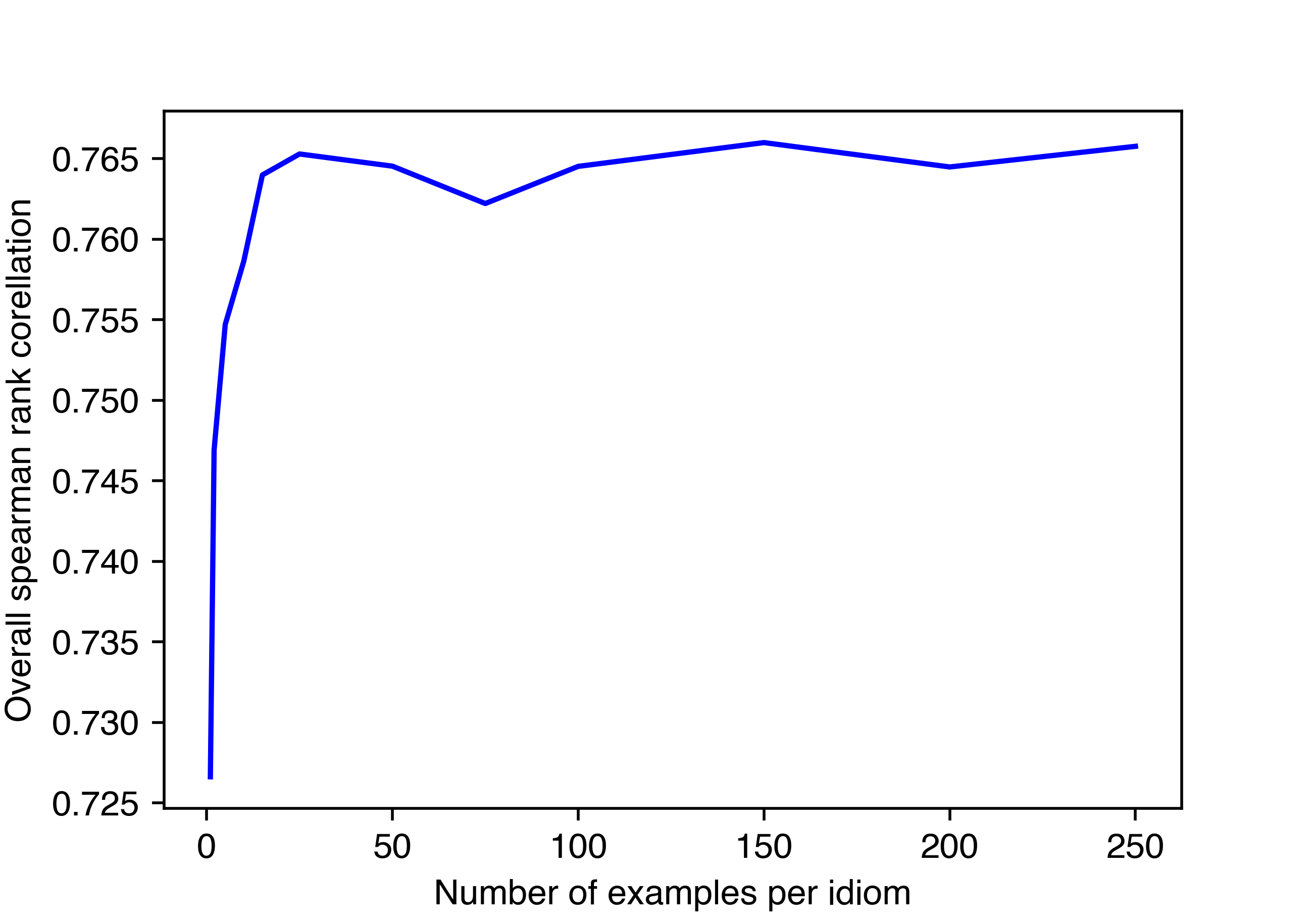}
   \caption{Overall Spearman Rank corellation score on the development set with different numbers of examples used to create the idiom embeddings.}
  \label{number-examples}
\end{figure}

We also tune the number of examples given for each idiom on the development data. Using BERTRAM we train embeddings for each of the idioms using a range of different numbers of examples from 1-250. The performance of each set of embeddings is evaluated by training the whole system for 10 epochs followed by evaluation on the dev set. Figure 3 shows the results of this experiment. The performance increases quickly from 1-15 examples before flattening out. The absolute highest performance is achieved at 150 examples, and so this is the value we use going forward.


\section{Results}

The final results for our system on the test data can be seen in Table \ref{final-results}. These scores show significant improvement over the baseline system and led to our system being placed first for the pre-train setting, and second for the fine-tune setting.

Fine-tuning has a much lower effect on the performance of the system when evaluated on the test set than compared with the dev and evaluation sets, with only a small, but significant, rise in overall correlation. Performance rises by only 0.0198 and 0.022 for English and Portuguese respectively, and unlike on dev data we do not see a uniform increase on the SR Idiom score.

\subsection{Galician Performance}

The performance we achieve on the Galician idiom data is much lower than what is seen on the English and Portuguese data. As we didn't have access to any development data for Galician further investigation will be needed to identify the causes of this discrepancy. Due to the smaller amount of Galician data in the cc100 corpus, some idioms did not have the full 150 examples that were used to create the embeddings for the English and Portuguese idioms. Additionally, there was no Galician STS data to train the final model on, and even though Portuguese and Galician are very similar, the small difference may lead to differences in the performance.




%
%
%


\subsection{Error Analysis and Data Issues}

To perform analysis on the quality of the created representations we calculate the Spearman's Rank Correlation for each of the idioms in the development set individually. Any idioms with less than 5 occurrences in the development data are removed, as significant correlation scores cannot be achieved with such a low sample size.

When evaluating the performance of the idioms individually, we can see that some of the idiomatic expressions perform much worse than average. For example the spearman rank for score for `fish story' is just 0.190 when the embedding is trained on 10 random examples. 

Analysis of these errors shows that the lower performance can, at least in part, be attributed to different phrase senses in the automatically collected examples. Taking our above example `\textit{fish story}', 3 different phrase senses can be observed in the original randomly selected examples: a tall tale, a literal story about fish, and as a proper noun in the title of the film `A Fish Story'. This leads to a divergence in the contexts in the examples, and the contexts for the idiomatic uses, leading to worse embeddings for the idiomatic phrases.

We can explore this further by producing a manually collected gold standard example set, for the English language subset of the MWEs. Taking the original 250 examples for each idiom, we select 10 gold standard examples. To avoid overfitting our embeddings to this task, we only manually remove examples where the MWE is being used as a proper noun (e.g. the film 'A Fish Story'), or the idiom is being misused, leaving in correct literal and idiomatic uses of the phrase. After removing the proper noun and misused cases, 10 random examples are selected to form our 'gold standard' example set.

We then compare the spearman scores achieved when the embeddings are trained with the gold standard examples, to scores when the representations are produced using 10 random examples when both models are evaluated on the English split of development set. The results for selected MWEs with the randomly selected (auto) and manually chosen (manual) contexts can be seen in table \ref{tab:manual}. 

The manually selected examples lead to an increase in performance on the  Idiom STS data split from 0.406 to 0.450. A small increase from 0.841 to 0.848 overall on the English split can also be observed, however this performance is limited by the general STS score which is unaffected by our manual selection. Particularly large improvements in spearman rank coefficient can be seen on MWEs with multiple meanings (panda car, banana republic, fish story, etc.). Surprisingly, we actually see the performance on some MWEs fall, however this can likely be attributed to the random selection of examples, and variance in the contexts used for each idiom, especially on the MWEs which did not have many usages removed as they are only used in the idiomatic form (eager beaver, chain reaction, etc.).

\begin{table}
\centering
\begin{tabular}{cccc}
\hline
\textbf{MWE} & \textbf{Auto} & \textbf{Manual} & \textbf{Change}\\
\hline
panda car & 0.399 & 0.851 & 0.452 \\
banana republic & 0.391 & 0.753 & 0.362 \\
... & ... & ... & ... \\
fish story & 0.190 & 0.304 & 0.114 \\
... & ... & ... & ... \\
chain reaction & 0.356 & 0.240 & -0.116 \\
eager beaver & 0.491 & 0.352 & -0.159 \\
\end{tabular}
\caption{Improvement in correlation, measured using Spearman's Rank Coefficient, when trained on manually chosen examples vs. automatically collected ones.}
\label{tab:manual}
\end{table}

\section{Conclusion}

We build our system by augmenting BERT models for each language with single token embeddings learnt using BERTRAM. BERTRAM is used due to its high performance on rare words, which share many properties with idioms such as non-compositionality and being rare examples of component pieces. Our results, and subsequent ranking at first place (of six entries) in the pre-train setting and second place (of five entries) in the fine-tune setting, show that BERTRAM can learn high-quality word embeddings for idioms and that this leads to better performance on downstream tasks. Our error analysis shows that BERTRAM is sensitive to the quality of examples it is shown, and that performance can be improved even further by manually selecting a gold set of contexts for each idiom. Future work could look at the differences in performance between the Portuguese and Galician models with the goal of increasing performance on Galician, and perform more analysis to explore the discrepancy in performance between individual idioms further.

\section*{Acknowledgements}

This work was supported the Healthy Lifespan Institute (HELSI) at The University of Sheffield and is funded by the Engineering and Physical Sciences Research Council [grant number EP/T517835/1].

\bibliography{anthology,custom}
\bibliographystyle{acl_natbib}

\label{sec:appendix}

\end{document}